\documentclass[lettersize,journal]{IEEEtran}
\usepackage{amsmath,amsfonts}
\usepackage{algorithm}
\usepackage{algpseudocode}
\usepackage{array}
\usepackage[caption=false,font=normalsize,labelfont=sf,textfont=sf]{subfig}
\usepackage{textcomp}
\usepackage{stfloats}
\usepackage{url}
\usepackage{verbatim}
\usepackage{graphicx}
\usepackage{cite}

\usepackage{booktabs}

\usepackage{enumitem}

\usepackage{xcolor}

\newcommand{\tabincell}[2]{\begin{tabular}{@{}#1@{}}#2\end{tabular}}

\begin{document}

\title{Urban Power Grid Topology and Hierarchy Identification from Open Data}

\author{Shiliang Zhang, Sabita Maharjan, Alexey Vinel, Shui Yu
\thanks{This work was supported by the PriTEM project funded by UiO:Energy Convergence Environments. Shiliang Zhang and Sabita Maharjan are with University of Oslo, Oslo, Norway (email: \{shilianz, sabita\}@ifi.uio.no). Alexey Vinel is with Karlsruhe Institute of Technology, Karlsruhe, Germany (email: alexey.vinel@kit.edu). Shui Yu is with University of Technology Sydney, Austrilia (email: shui.yu@uts.edu.au).}
}



\maketitle

\begin{abstract}
Understanding the complex topology and hierarchy of urban power grid is crucial for energy prognosis, power flow management, and system resilience analysis. However, detailed grid information remains largely proprietary. This information, nevertheless, is crucial especially for analyzing the last-mile distribution networks connecting individual buildings. This paper addresses this challenge by developing an open-data-driven framework to identify urban power grid topology, from high-voltage transmission down to individual building connections. We integrate public infrastructure data (power-lines, substations, transformers, poles) to map the high and medium-voltage grid skeleton. We then leverage geospatial machine learning on OpenStreetMap building data to cluster loads of building energy demand. Upon the clustering, we develop the physical topology of the final distribution lines linking the clustered buildings and low-voltage distribution grids. Our derived grid topology incorporates physical and geographical constraints like voltage level and road networks, thus leading to reasonable topology and hierarchy of grid networks. We apply the developed framework to the districts of Alna in Oslo, Norway, and Waiheke Island in New Zealand, and we reconstruct the complete grid topology that connects all the buildings, power lines, and electricity infrastructure assets. With the developed framework, we provide a critical tool that facilitates power system analysis, \textit{e.g.}, power flow optimization, cascading failure simulation, and grid resilience against the penetration of distributed renewable generation. This work also opens the opportunity to build energy digital twins that fuse power assets and grid topology, geographic constraints, and fine-grained power demand profiles.

\end{abstract}


\section{Introduction}
\IEEEPARstart{T}{he} urban power grid serves as the backbone for modern society, which ensures consistent supply of electricity that underpins economic activities, public services, and daily life~\cite{9954283}. As the transition towards sustainable and resilient energy systems unfolds~\cite{10360247}, the power grid topology (the connectivity pattern of power nodes and lines) and hierarchy (the organization across grid infrastructure types) have become increasingly important~\cite{alhazmi2025advancing}. Granular models of these structures are indispensable for critical energy applications, including power flow optimization, energy prognosis, and simulation of power system resilience.

Neverthelss, detailed information about urban power infrastructure remains largely proprietary~\cite{doi.org/10.1049/stg2.12161,sturmer2024increasing,10.1145/3639036}. Utility agencies may hold their distribution network map as confidential, due to safety \& security concerns and commercial constrains~\cite{11015807}. Though high-voltage transmission skeletons are often partially available or inferable~\cite{xiong2025modelling,9374103}, the lack of transparency is a big challenge in distribution network applications~\cite{9582801}, particularly the last-mile connections that link transformers, utility poles, and individual end-user buildings. This gap hinders the development of models necessary for analyzing local energy phenomena, such as the impact of distributed renewable penetration \textit{e.g.} from rooftop solar panels, and the simulation of cascading failures within microgrids~\cite{10.1145/3679240.3734602,10336916,10547711,10066154}. It also impedes the studies on urban energy transition, where battery storage, electric vehicles, and demand-response schemes can materially alter local demands and power flows~\cite{BIRKJONES20228225,10538433,VANKUDOTH2024112691}.

Existing studies in bridging this gap often rely on either synthetic models generated through simplified geometric assumptions, or synthesized data that lacks geographic and physical network basis. Jongh \textit{et~al.}~\cite{10742905} considered the distribution grid of fixed structure yet with the active status of its lines unknown, and they realized the active status of cables with limited measurements. Li \textit{et~al.}~\cite{LI2025111649} targeted a grid with known structure, and detected the grid topology change due to the operations of soft open points (SOPs) and unreported deactivated lines. Bolognani \textit{et~al.}~\cite{BOLOGNANI2024109} assumed a first-order model for the distributed grids, and recognized the topology with grid measurements. Huang \textit{et~al.}~\cite{HUANG2024110591} considered the power lines linking transformers and customers, and proposed a Gaussian mixture model in identifying the grid topology in an adaptive way. Luan \textit{et~al.}~\cite{https://doi.org/10.1049/esi2.12142} realized transformer-customer pairing in distribution network using K-nearest neighbor (KNN) approach. Ahshan \textit{et~al.}~\cite{AHSHAN2025100486} introduced a convolutional-network based approach to locate the grid infrastructure in distribution power grid, \textit{e.g.}, substations and poles, without identifying power lines connecting them. Tischbein \textit{et~al.}~\cite{10738069} used synthesized power flow data to train an LSTM model to determine whether a link exists between two given nodes in the grid, while neglecting geographical constraints for the link. Feng \textit{et~al.}~\cite{10612997} assumed that the network topology satisfies a distance topology matrix (DTM) structure. Based on simulated meter data, they seek the most representative DTM for low-voltage distribution grid, and reconstruct the grid topology through clustering approach. 

There are also studies identifying grid topology from real-time meter readings. Karunarathne \textit{et~al.}~\cite{10705904} identified low-voltage network topology through regression and clustering, using real-time high-resolution smart meter data. Zhu \textit{et~al.}~\cite{https://doi.org/10.1049/gtd2.70064} identified the grid topology using both real-time and historical measurement data. Xu \textit{et~al.}~\cite{9944182} identified distributed network topology using smart metering data, where they considered the scenarios that multiple feeders are connected with the same bus. A similar metering-based approach is presented by Guo \textit{et~al.}~\cite{11063399}, where they further developed the voltage quality control for the identified grid topology. Nozal \textit{et~al.}~\cite{RODRIGUEZDELNOZAL2024109184} described a genetic approach in modeling distribution grid topology using metering data. Yang \textit{et~al.}~\cite{yang2024edge} employed principal component analysis on metering data to gain topological connections between nodes in distribution grid. Li \textit{et~al.}~\cite{10518183} proposed a weighted convolution model to identify transformer-customer connectivity. Their identification relies on analyzing measured power data from transformers, customers, and household PV systems. Zheng \textit{et~al.}~\cite{10693131} adopted graph theory to represent the grid topology, and used optimization techniques in the topology recognition. Huang \textit{et~al.}~\cite{huang2024automatic} recognized the grid topology of station area, using Lasso algorithm and t-SNE algorithm, while they did not consider the grid hierarchy across station areas. 

Investigations haven been explored to reduce the usage of real metering data. García \textit{et~al.}~\cite{GARCIA2025111517} presented topology identification for low-voltage distribution networks based on wavelet transform. Their method requires less amount of data and grid observability for the identification. Ponsaerts \textit{et~al.}~\cite{11180674} developed a topology identification combining admittance matrix estimation and hidden nodes detection, with limited smart meter coverage in the grid. Shi \textit{et~al.}~\cite{10808491} identified station area grid topology with part of metering data missing, where they compensate the missing values and use K-mean clustering to process the data for the topology identification. Tischbein \textit{et~al.}~\cite{10863762} analyzed the impact of measurement uncertainties (\textit{e.g.}, missing or incorrect metering data) on grid topology identification. Jongh \textit{et~al.}~\cite{10863283} presented a topology identification through mixed-integer quadratic programming on power usage measurements and geographic information system (GIS) data, where they deduct grid topology based on spatial relationships between measurements. Nevertheless, collecting meter readings from individual buildings at district or city level challenges privacy requirement~\cite{HEUNINCKX2023103040}, rendering it complicated in implementing meter-reading-based network topology identification at large scales. 

Over the recent years, the open data ecosystem has advanced rapidly that can release privacy constraints. Public geospatial datasets provide rich coverage of urban power infrastructure footprints, including overhead and underground power lines, substations, transformers, \textit{etc}~\cite{zhang2025norwegian,ENGAN2025111453,NEURIPS2024_c7caf017}. Open Street Map (OSM) offers building-level geometries with attributes indicating energy demand~\cite{f2024openstreetmap}, and authoritative agencies publish data on utilities and power consumption and production. Advances in graph analytics and geospatial machine learning~\cite{ghosh2024graph} make it feasible to fuse these disparate data into coherent infrastructure network models. 

Topology identification from open data has been explored in recent studies. Weber \textit{et~al.}~\cite{10540632} introduced a semi open-data-driven approach in distribution network identification. Their identification uses the open data of OSM building footprints and 3D OSM building data, together with proprietary data of the number and location of smart meters from the DSO, with the transformer location assumed to be known. Liu \textit{et~al.}~\cite{LIU2025110395} leveraged open GIS data and smart meter data to identify the distribution network topology. They used GIS data in building the topology, and then trimmed the topology branches and improved the network representativeness using smart meter data. 

We also note open-data-based approaches without utilizing smart meter data. Baecker \textit{et~al.}~\cite{REVERONBAECKER2025101617} presented an open-source approach where they synthesized distribution grid topology. They refered to open and georeferenced data for buildings, streets, and transformers, and generated a representative topology for low-voltage power grid. Lundblad \textit{et~al.}~\cite{LUNDBLAD2024101483} developed an open approach for synthesizing low-voltage grid at a national level, where they emphasized the accuracy of the synthesized demand and the grid capacity. Oneto \textit{et~al.}~\cite{ONETO2025101678} introduced large-scale generation of synthetic georeferenced distribution grids using open data, taking into account operational and geographical constraints. Xiong \textit{et al.}~\cite{xiong2025modelling} reconstructed the European high-voltage grid using OpenStreetMap data. They included components of substations, transmission lines and cables, transformers, and converters in the topology identification. Similarly, Wang \textit{et~al.}~\cite{wang2023geospatial} reconstructed distribution networks using open data. Their work leverages multi-modal data like road network, building maps, and street view images, and reconstruct overhead and underground grid topology. Nevertheless, the above studies did not consider low-voltage infrastructure and the ``last-mile'' power lines linking individual buildings. Furthermore, the hierarchy of the network is largely not included in the identification, rendering that the resulted grid network cannot really contribute to advanced studies like power flow analysis, optimization, and prognosis. These mentioned developments together with their limitations motivate us in identifying urban power grid topology and hierarchy from open data, such that it can reflect both the skeleton and the ``last-miles'' of the grid in a realistic manner.

In this paper, we present an open-data-driven framework for the complete, bottom-up identification of urban power grid topology and hierarchy. Our method utilizes open-source geospatial data to reconstruct the network from high-voltage power lines down to individual building connections. Particularly, we identify the high- and medium-voltage skeleton using graph-based method, while respecting voltage levels, asset types, and typical meshed patterns observed in the urban context. We then generate building-level demand clusters by using geospatial machine learning and the data from OSM. We establish the power line connections among clustered buildings using the minimum spanning tree (MST) approach~\cite{9747921}. Finally, we infer the complete power grid network that links all the individual buildings, building clusters, poles, transformers, substations, and high- and medium-voltage distribution power lines. We validate and demonstrate the developed framework in the district of Alna, Oslo, Norway, a heterogeneous urban area with a mix of residential, commercial, and industrial loads. Using publicly available infrastructure and OSM building footprints, we reconstruct the connectivity for 7,330 buildings, alongside high- and medium-voltage power lines, substations, transformers, and poles that traverse the district. We further validate our approach in Waiheke Island in New Zealand, which has significantly distinct grid network features compared to Alna. This demonstrates that our approach is robust and works across diverse grid network scenarios.

We summarize our contributions in this paper as follows:

\begin{enumerate}[label=(\roman*)]
    \item We present an open-data-driven framework in identifying urban power grid topology and hierarchy, where we integrate data including power lines, substations, transformers, utility poles, and building footprints. This leads to a network topology identification that incorporates physical and geographical constrains.
    \item We develop the identification of grid skeleton leveraging graph-based algorithm, OSM data, and publicly available infrastructure data. This results in the network topology that links high- and medium-voltage lines, substations, transformers, and poles.
    \item Through clustering urban buildings and geospatial machine learning on OSM building footprints, we infer the power links connecting individual buildings in clusters. In this way, we establish the ``last-miles'' for the distribution network in the topology identification. 
\end{enumerate}

Through this effort, we reconstruct the complete network topology from high-voltage lines down to building demands. We apply and validate our approach in two areas with distinct power infrastructure and geographical attributes, and demonstrate the effectiveness of our method across diverse grid network features.

\begin{figure*}[tbhp]
    \centering
    \includegraphics[width=0.99\linewidth]{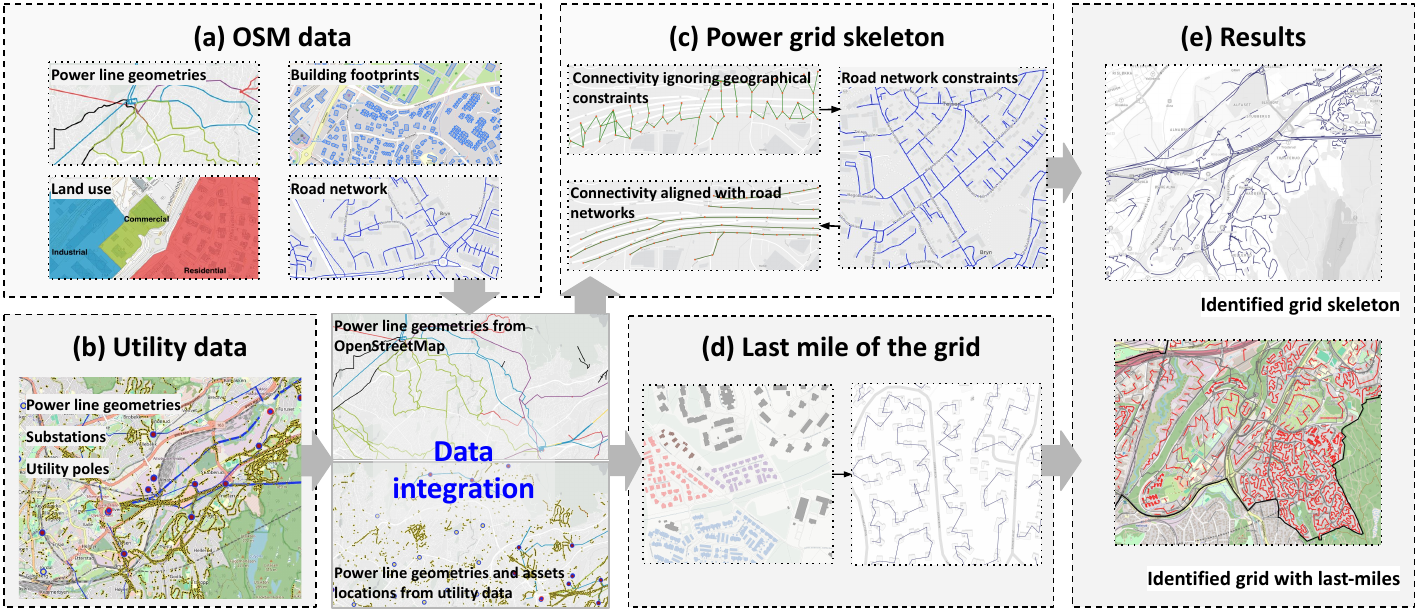}
    \caption{Overview of the workflow for our urban power grid identification. (a) presents the data from OpenStreetMap including power line geometries, land-use type of areas that urban buildings belong to, footprints of individual buildings, and road network topology. (b) exemplifies the data openly released by governmental agencies (NVE in this case), encompassing the data of power line geometries and locations of power infrastructure. (c) demonstrates the grid skeleton topology that is aligned with geographical constraints. (d) displays the ``last-miles'' of the power grid linking individual buildings. (e) shows the outcome of topology reconstruction integrating the grid skeleton and the last-miles.}
    \label{fig:workflow_overview}
\end{figure*}

The reminder of this paper is as follows. Section~\ref{subsec:overview} provides an overview of our open-data-driven topology identification framework. Section~\ref{subsec:skeleton} describes how we reconstruct the power grid skeleton, encompassing high-, medium-, and low-voltage power lines, substations, transformers, and poles. Section~\ref{subsec:last_mile} present the reconstruction of grid lines connecting buildings, which establishes the ``last-mile'' of the urban power grid in completing the network identification. Section~\ref{sec:case_study} applies and demonstrates our framework in the district of Alna, Oslo, Norway. Section~\ref{sec:further_case_study} further validates our framework in Waiheke Island, New Zealand, which significantly distinguishes from the grid network features of Alna. We discuss the implications of releasing distribution network topology with the last-miles in Section~\ref{sec:implication}, and we conclude this paper in Section~\ref{sec:conclusion}.

\section{Urban power grid topology identification}\label{sec:method}

\subsection{Overview of our approach}\label{subsec:overview}

Fig.~\ref{fig:workflow_overview} provides an overview of our approach encompassing different components. We start the grid topology identification by collecting relevant data that includes utility data openly released by authoritative agencies (\textit{e.g.}, NVE\footnote{\url{https://kartkatalog.nve.no}, accessed 2026-09-01.}) and OSM data. These two data resources are complementary, and we integrate them for the entire topology reconstruction. We use the integrated data to generate the grid skeleton, where we build the power lines connecting high-, medium-, and low-voltage power lines and major power infrastructure including substations, transformers, and utility poles. We generate the connectivity respecting operational and geographical constraints, as exemplified in Fig.~\ref{fig:workflow_overview} (c). We then construct the ``last-miles'' of the grid with the power lines linking all the individual buildings and the power grid skeleton. In generating the last-miles, we partition the area of interest into sub-areas, within each of which we build the connections. The area partition is conducted according to the land-use type (\textit{e.g.}, residential or industrial), and the power line links are generated through geographic machine learning. Below we elaborate on identification of the grid skeleton and the last-miles.

\subsection{Skeleton of the urban power grid}\label{subsec:skeleton}

In this section, we describe how we reconstruct the skeleton of the urban distribution power grid topology. We refer to this skeleton as the hierarchical graph encompassing all major infrastructure assets (substations, transformers, power lines) with all known and inferred connections across high-voltage, medium-voltage, and low-voltage levels. This skeleton topology forms the foundation upon which the ``last-mile'' connectivity to individual buildings will be established in Section~\ref{subsec:last_mile}.\\

\subsubsection{Data integration}\label{subsec:data_integration}

The skeleton identification relies on the synergy of two open data sources: (i) OpenStreetMap (OSM) and (ii) utility data openly released by authoritative agencies, \textit{e.g.}, NVE (Norwegian Water Resources and Energy Directorate) in our work.

The OSM data retrieved includes (i) land-use label and boundaries for urban areas and (ii) the road network geometry. Particularly, the former is used to partition the urban area. In this way, we can cluster buildings and power infrastructure into groups, and identify the topology in each groups in parallel. The latter is used to guide topology reconstruction of low-voltage infrastructure like utility poles.

\begin{figure}
    \centering
    \includegraphics[width=1\linewidth]{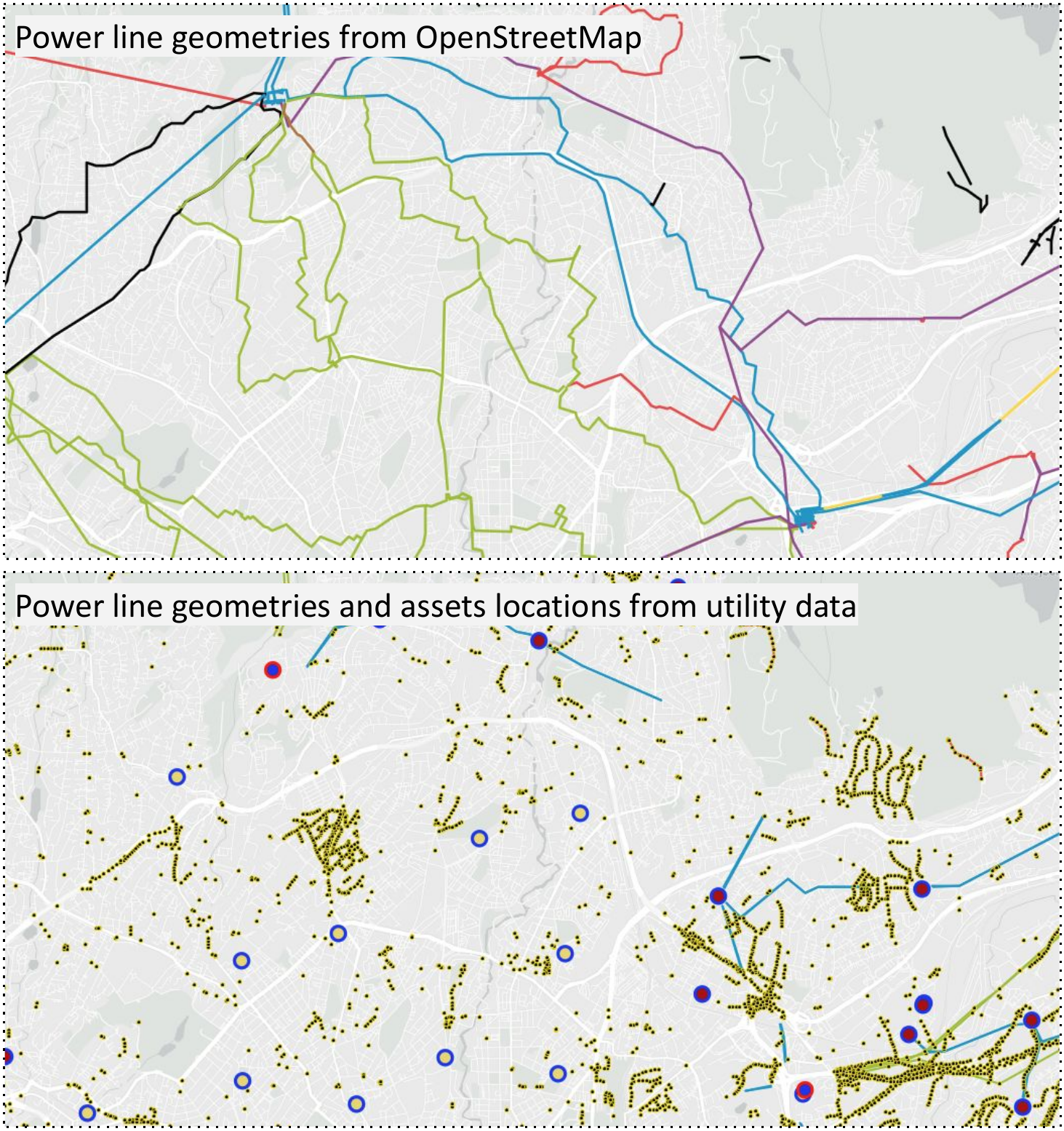}
    \caption{An example of data sources from OSM (upper) and utility data released by authoritative agency NVE (lower). The lines are transmission or distribution power grid, and the dots in the lower are utility poles (black solid dots with yellow edge) and substations (bigger and hollow dots).}
    \label{fig:Data_source}
\end{figure}

The utility data provides geospatial information for the physical assets, including location of substations, transformers, utility poles, along with the line geometry for transmission and distribution lines. Note that we also use line geometry from OSM to compensate the utility data where there are missing values, as exemplified in Fig.~\ref{fig:Data_source}.

We combine the two data sources to categorize assets based on voltage level and the hierarchical structure of the grid, as shown in Table~\ref{tab:assets}, and generate the topology for the skeleton of the grid.\\

\begin{table}
    \centering
    \caption{Categorized assets in urban grid and their connectivity basis}
    \resizebox{0.48\textwidth}{!}{
    \begin{tabular}{ccc}
        \hline
        Voltage level & Asset involved & Connectivity basis\\
        \hline
        \tabincell{c}{High- and\\ medium-voltage} & \tabincell{c}{Transmission lines,\\ substations, transformers,\\distribution lines} & \tabincell{c}{Merged line geometries\\ from utility data and OSM}\\
        \hline
        Low-voltage & Utility poles & \tabincell{c}{Inferred links constrained\\ by road geography}\\
        \hline
    \end{tabular}}
    \label{tab:assets}
\end{table}

\subsubsection{High- and medium-voltage network reconstruction}

This reconstruction concerns the connections between transmission and distribution power lines, together with the grid network's connection with substations and transformers. All the geometries/locations of the power lines and assets are retrieved from OSM or utility data.

\begin{figure}[tbhp]
    \centering
    \includegraphics[width=0.8\linewidth]{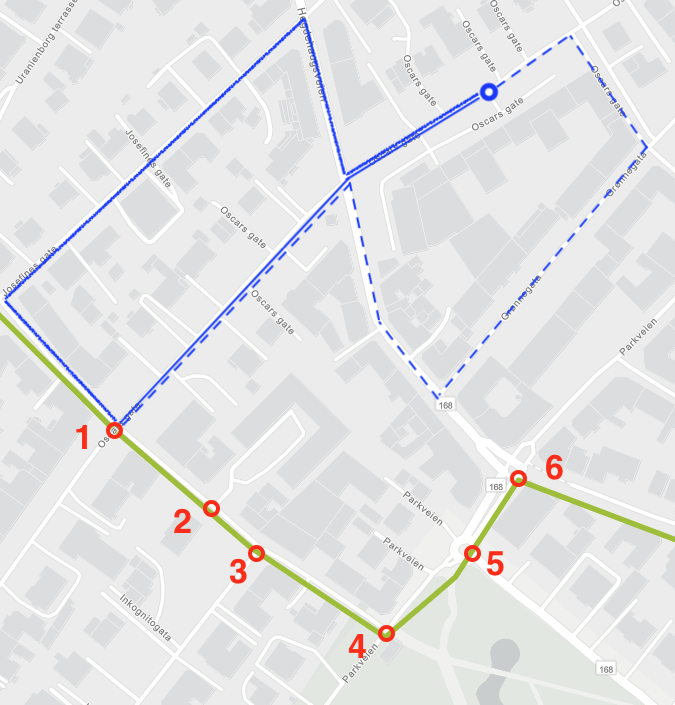}
    \caption{An example of connecting an asset with a power line. The green line represents a power line, the red dots are the potential points, the blue dot is the asset. We display three possible connections between the asset and potential point 1, and the shortest connection is selected through Dijkstra's algorithm. This connection recognition repeats for point 2-6. The final connection between the asset and the power line is determined as the shortest connection.}
    \label{fig:links}
\end{figure}

We connect power lines in a way that respects power flow operations. That is, the electricity flows from lines of higher voltage to lower uni-directionally, and from transmission lines to distribution lines. The substations/transformers should also match the voltage level of the power lines they are connected to. We also assume that all the power line connections respect the geographical constraints of the road network. Fulfilling the above constraints, we develop a graph-based approach to reconstruct the grid topology. Particularly, for each of the power line, we first generate the potential points on the power line that can be linked to a target power grid asset. We refer to potential point as the location on the power line that intersects a road or overlaps with an intersection in the road map. Then we calculate the shortest connection between those potential points and the target asset using Dijkstra's algorithm~\cite{9397200}, as shown in Fig.~\ref{fig:links}. The full procedure of generating the links complied with all the operational and geographical constraints is explained in the pseudocode Algorithm~\ref{alg:1}

\begin{algorithm}
    \caption{High/Medium-voltage network reconstruction}\label{alg:1}
    \begin{algorithmic}[1]
        \State \textbf{Input:} GIS information of high- and medium-voltage power lines and assets
        \State \textbf{Output:} Connections between power lines and substations/transformers
        
        \Procedure{}{}
            \State Group power lines by voltage level
            \State Sort groups in ascending order of voltage levels
            
            \For{each voltage level group $i$}
                \For{each power line in group $i$}
                    \State Generate potential connection points for lower voltage power lines
                    \State Identify the next higher voltage group $i+1$
                    \For{each power line in group $i+1$}
                        \State Generate potential connection points
                        \State Use Dijkstra's algorithm to find shortest path between power lines from group $i$ and group $i+1$ through potential points
                        \State Store the connection details
                    \EndFor
                \EndFor
            \EndFor
            
            \For{each power line}
                \If{the power line connects to a transformer}
                    \If{transformer voltage level matches the power line}
                        \State Create a direct link between power line and transformer, by generating potential points on the power line and realizing the link using Dijktra's algorithm
                    \EndIf
                \EndIf
            \EndFor
            
        \EndProcedure
    \end{algorithmic}
\end{algorithm}

\subsubsection{Low-voltage assets connectivity inference} This section concerns with the power grid connections between utility poles. To reconstruct the grid topology, we employ a hierarchical clustering approach combined with geographical constraints derived from OSM road network to model the connectivity. Below we elaborate the reconstruction in detail.

The reconstruction starts with a two-stage spatial partition of the utility poles in urban area. We first divide the considered area into sub-areas based on land-use boundaries (\textit{e.g.}, residential, commercial, industrial) available in OSM. This ensures that the resulted topology is consistent with geographically distinct land-use zones, aligning with typical urban planning and utility deployment strategies.

We then conduct an intra-area clustering. That is, within each sub-area, we cluster the utility poles using hierarchical density-based spatial clustering of applications with noise (HDBSCAN)~\cite{Fuchs2022}. This algorithm is selected as it can identify pole clusters of varying density and shape.

\begin{figure}[tbhp]
    \centering
    \includegraphics[width=0.9\linewidth]{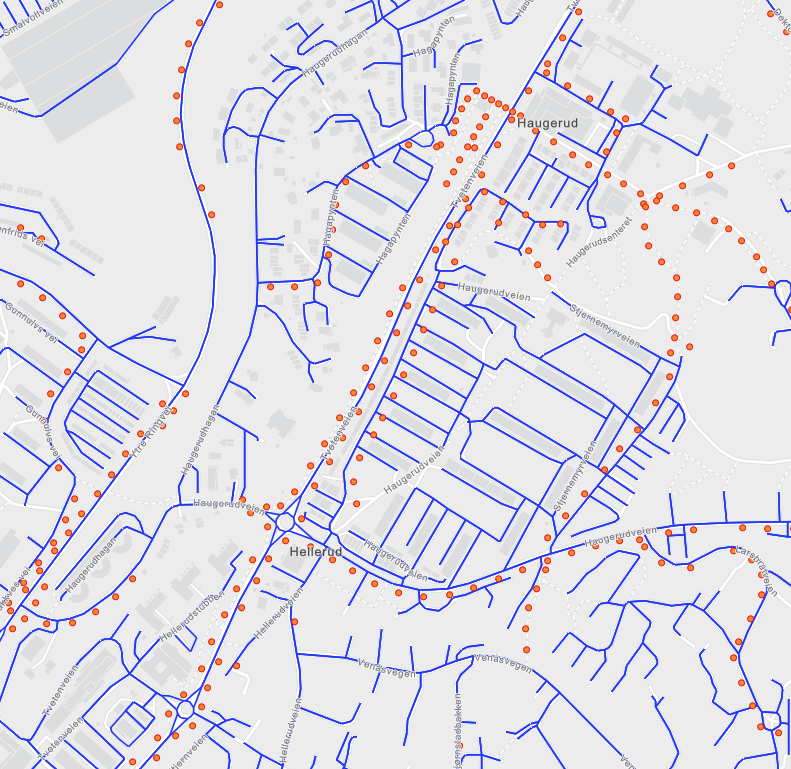}
    \caption{An example of the distribution of utility poles (red dots). The blue lines are the main road network.}
    \label{fig:utility_poles}
\end{figure}

Upon the spatial partition, we establish the connections between utility poles by generating edges within the identified clusters, prioritizing connections that respect real-world utility installation practice along the road network, as demonstrated in Fig.~\ref{fig:utility_poles}. This process ensures the resulted topology is geometrically plausible.

Particularly, for every utility pole in a cluster, we prioritize its connection to a neighboring pole based on two constraints:

\begin{enumerate}[label=(\roman*)]
    \item The connection should be located on the same side of the nearest road segment. We verify this condition by projecting the two poles onto the road segment and checking the direction of the offset vectors.
    \item The two poles should be near each other in the same cluster.
\end{enumerate}

If a pole cannot satisfy both constraints, then the constraints are relaxed and the pole is connected to its globally nearest utility pole within the same cluster. This fallback ensures full connectivity while maintaining a preference for the most probable topology.

We also note that HDBSCAN can lead to outliers of utility poles that are not assigned to any clusters. In that case, we link the outlier poles to their nearest utility pole regardless of cluster membership. We present the full procedure of reconstructing grid topology amongst utility poles in Algorithm~\ref{alg:lv_connectivity}

\begin{algorithm}
\caption{Utility pole connectivity inference via constrained HDBSCAN}
\label{alg:lv_connectivity}
\begin{algorithmic}[1]
    \renewcommand{\algorithmicrequire}{\textbf{Input:}}
    \renewcommand{\algorithmicensure}{\textbf{Output:}}
    
    \Require Utility pole locations $V_{Pole}$, land use boundaries $\mathcal{L}$, road network geometry $\mathcal{R}$
    \Ensure Utility pole edge set $E_{p}$

    \State Initialize the edge set: $E_{p} \leftarrow \emptyset$ 
    
    \noindent\textbf{Stage 1: Hierarchical spatial partitioning}. 
    
    \noindent // Partition urban area into sub-areas $A_k$ based on land use $\mathcal{L}$ 
    \State $\mathcal{A} \leftarrow \text{Partition}(\text{Area}, \mathcal{L})$ 
    
    \For{each sub-area $A_k \in \mathcal{A}$}

    // Find poles within the sub-area
        \State $V_{A_k} \leftarrow V_{Pole} \cap A_k$ 

        // Cluster poles using HDBSCAN
        
        // $V_{A_k}^{Clust}$ are the identified clusters $C_{k,j}$
        
        // $V_{A_k}^{Out}$ are the outlier poles
        \State $V_{A_k}^{Clust}, V_{A_k}^{Out} \leftarrow \text{HDBSCAN}(V_{A_k})$ 
        
        \noindent \textbf{Stage 2: Constrained link generation}
        \For{each cluster $C_{k,j} \in V_{A_k}^{Clust}$}
            \For{each pole $P_i \in C_{k,j}$}

            // Find all other poles in cluster

            // Apply two constraints
                \State $N_{C_{k,j}} \leftarrow \text{Neighbors}(P_i, C_{k,j})$ 
                

                // Find geometrically nearest pole
                \State $P_{near} \leftarrow \text{NearestNeighbor}(P_i, N_{C_{k,j}})$ 
                
                \If{$\text{IsSameSide}(P_i, P_{near}, \mathcal{R}) == \text{TRUE}$}

                // Connect if same side and nearest
                    \State $E_{p} \leftarrow E_{p} \cup \{(P_i, P_{near})\}$ 
                \Else

                // Fallback: Connect to nearest neighbor
                    \State $E_{p} \leftarrow E_{p} \cup \{(P_i, P_{near})\}$ 
                \EndIf
            \EndFor
        \EndFor
        
        \noindent \textbf{Stage 3: Outlier handling}
        \For{each outlier pole $P_o \in V_{A_k}^{Out}$}
        
        // Find nearest pole in entire $V_{Pole}$ set
            \State $P_{glob\_near} \leftarrow \text{NearestNeighbor}(P_o, V_{Pole})$ 
            \State $E_{p} \leftarrow E_{p} \cup \{(P_o, P_{glob\_near})\}$
        \EndFor
    \EndFor
    
    \Return $E_{p}$
\end{algorithmic}
\end{algorithm}

\subsection{The last-mile of urban distribution network}\label{subsec:last_mile}

The final stage of the grid topology reconstruction is the inference of the last-mile connectivity, which establishes the grid lines linking individual buildings to the utility poles. This step requires geospatial algorithms to group demand points (buildings) and connect them to the nearest feasible supply point while respecting infrastructural and geographical constraints. We elaborate on this stage in the following.

\subsubsection{Spatial partition and building clustering}
The process of inferring building connectivity is computationally intensive and highly dependent on local geographical context. To manage this complexity, we apply the similar hierarchical approach in Section~\ref{subsec:skeleton} and partition the buildings before generating the power links.

\begin{figure}[tbhp]
    \centering
    \includegraphics[width=0.9\linewidth]{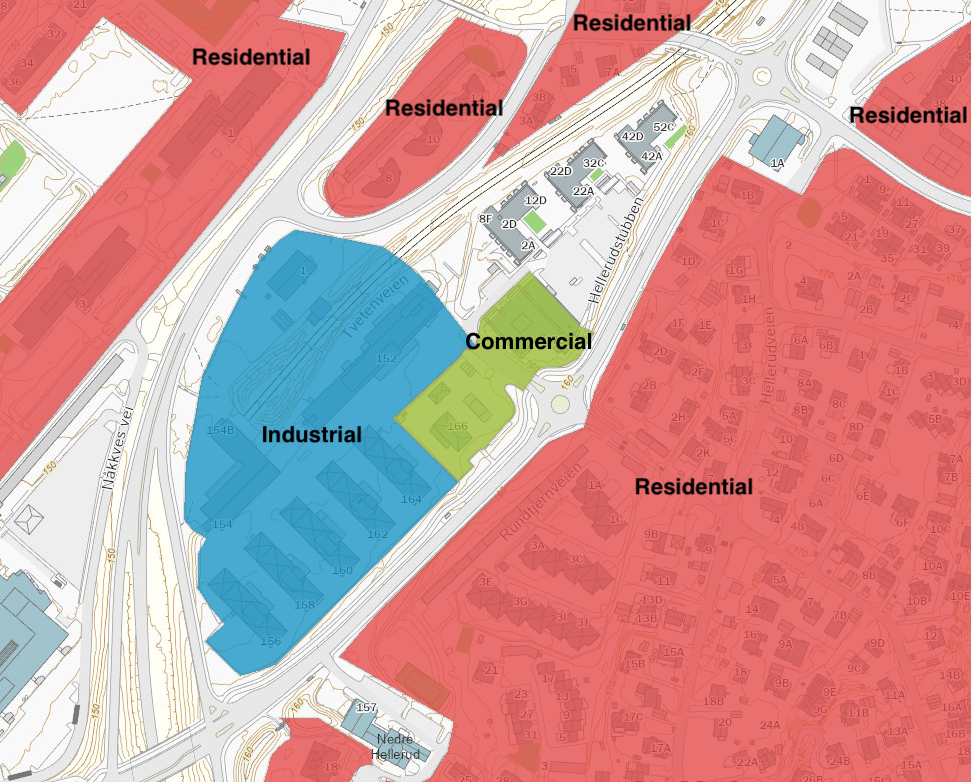}
    \caption{An example of land-use types retrieved from OSM.}
    \label{fig:land-use}
\end{figure}

We partition the entire study area into sub-areas according to the land-use types retrieved from OSM, as shown in Fig.~\ref{fig:land-use}, followed by creating power lines linking buildings within each sub-area. This ensures that the area partition reflect typical building densities, and that the consequent building power line connections reflect local urban environment.

\begin{figure}[tbhp]
    \centering
    \includegraphics[width=0.9\linewidth]{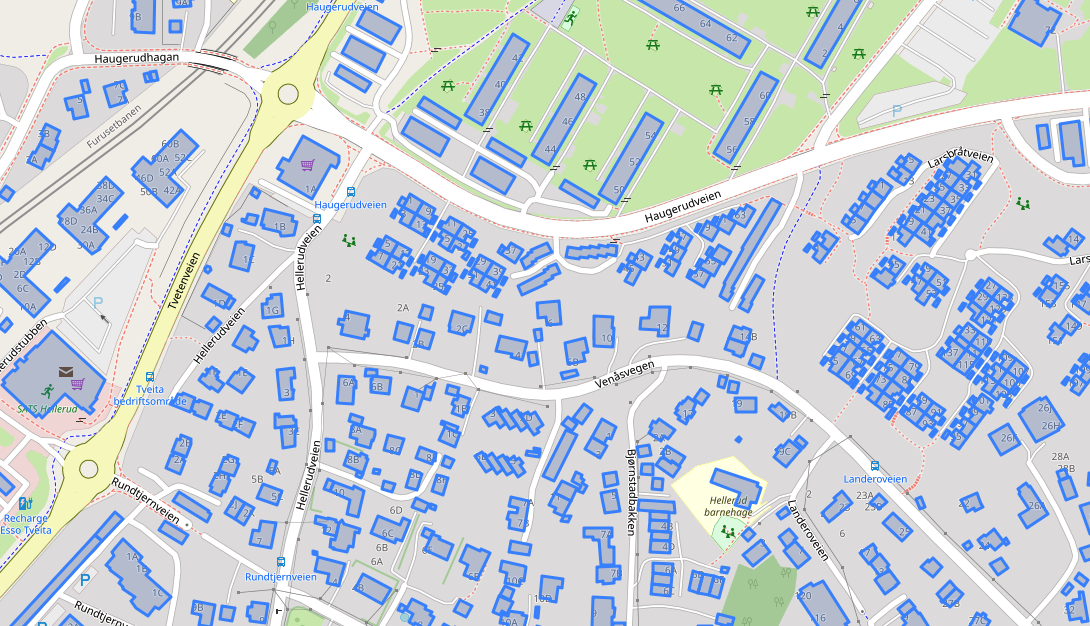}
    \caption{An example of building footprints (in blue polygons) retrieved from OSM.}
    \label{fig:building-footprints}
\end{figure}

Then within each sub-area, we cluster the georeferenced locations of the building footprints, as shown in Fig.~\ref{fig:building-footprints}, using HDBSCAN. This step identifies natural group of buildings that are likely to be served by a common utility pole or a supply point in a distribution line. The HDBSCAN output yields both clustered buildings and outlier buildings that do not belong to any clusters.

\subsubsection{Intra-cluster building connectivity}
For each of the clusters in a sub-area, we generate the connections that link all the buildings within a cluster while respecting geographical constraints. Particularly, we apply the minimum spanning tree (MST) algorithm to the building footprints, thus we can obtain the power line links for buildings within a cluster.

Note that there can be buildings not belonging to any clusters, In that case, we connect such a building to its nearest clustered building in the same sub-area. We generate such a connection using Dijkstra's algorithm that respects road network topology.

\subsubsection{Cluster-to-skeleton linkage}

The final step is to connect the building clusters to the reconstructed grid skeleton (see skeleton grid reconstruction in Section~\ref{subsec:skeleton}). This step specifically concerns the grid connection between building clusters and lowest voltage distribution assets, \textit{e.g.}, utility poles and low-voltage lines.

For each building cluster, the optimal connection point is determined by comparing two network-constrained shortest distances: (i) distance to the nearest utility pole $D_p$, and (ii) distance to the nearest low-voltage power line $D_l$. $D_p$ is calculated as the shortest path from any building in a cluster to any utility pole in the considered area, via the road network. $D_l$ is calculated as the shortest path from any building in a cluster to the closest point on any of the distribution grid with lowest voltage level. This accounts for scenarios where a cluster is optimally connected to an existing line segment rather than a utility pole.

\begin{algorithm}
\caption{Last-mile connectivity inference}
\label{alg:last_mile_connectivity}
\begin{algorithmic}[3]
    \renewcommand{\algorithmicrequire}{\textbf{Input:}}
    \renewcommand{\algorithmicensure}{\textbf{Output:}}
    
    \Require Building locations $V_{Bld}$, land-use boundaries $\mathcal{L}$, road network graph $\mathcal{R}$, utility poles $V_{Pole}$, low-voltage power lines $L_{LV}$

    \Ensure Last-mile edge set $E_{Last}$~\\~\\

    \noindent // Initialize the set of all last-mile edges

    \State $E_{Last} \leftarrow \emptyset$ 
    
    \noindent\textbf{Stage 1: Building clustering and partitioning}

    \noindent // Partition area into sub-areas $A_k$ based on land-use $\mathcal{L}$
    
    \State $\mathcal{A} \leftarrow \text{Partition}(\text{Area}, \mathcal{L})$ 
    
    \For{each sub-area $A_k \in \mathcal{A}$}

     // Buildings within the sub-area
        \State $V_{A_k}^{Bld} \leftarrow V_{Bld} \cap A_k$ 

        // Cluster buildings using HDBSCAN: $V_{C}^{k}$ is set of clusters, $V_{O}^{k}$ are outliers
        \State $V_{C}^{k}, V_{O}^{k} \leftarrow \text{HDBSCAN}(V_{A_k}^{Bld})$ 

        // Set of all clustered buildings
        \State $V_{Clustered}^{k} \leftarrow \bigcup C_{k,j} \text{ for } C_{k,j} \in V_{C}^{k}$ 

        \noindent\textbf{Stage 2: Intra-cluster connectivity by MST}
        \For{each cluster $C_{k,j} \in V_{C}^{k}$}
            \For{each pair of buildings $(B_i, B_j) \in C_{k,j}$}

            // Network-constrained distance using Dijkstra
                \State $w_{ij} \leftarrow \text{ShortestPath}(B_i, B_j, \mathcal{R})$ 
            \EndFor

            // Apply minimum spanning tree (MST)
            \State $E_{MST} \leftarrow \text{MST}(C_{k,j}, w_{ij})$ 
            \State $E_{Last} \leftarrow E_{Last} \cup E_{MST}$
        \EndFor

        \noindent\textbf{Stage 3: Outlier building connection}
        \For{each outlier building $B_o \in V_{O}^{k}$}

        // Find nearest clustered building
            \State $B_{near} \leftarrow \underset{B_c \in V_{Clustered}^{k}}{\operatorname{argmin}} (\text{ShortestPath}(B_o, B_c, \mathcal{R}))$ 

            // Add link following road network
            \State $E_{Last} \leftarrow E_{Last} \cup \text{Path}(B_o, B_{near}, \mathcal{R})$ 
        \EndFor
        
        \noindent\textbf{Stage 4: Cluster-to-skeleton linkage}
        \For{each cluster $C_{k,j} \in V_{C}^{k}$}
        
            // Distance to nearest pole $D_p$
            \State $D_p, (B_p^*, P^*)=$ 
            
            $\underset{B_i \in C_{k,j}, P \in V_{Pole}}{\operatorname{argmin}} (\text{ShortestPath}(B_i, P, \mathcal{R}))$
            
            // Distance to nearest low-voltage line segment $D_l$
            \State $D_l, (B_l^*, P_l^*)=$ 
            
            $\underset{B_i \in C_{k,j}, P_l \in L_{LV}}{\operatorname{argmin}} (\text{ShortestPath}(B_i, P_l, \mathcal{R}))$
            
            \If{$D_p \le D_l$}

            // Link cluster to pole
                \State $E_{Last} \leftarrow E_{Last} \cup \text{Path}(B_p^*, P^*, \mathcal{R})$ 
            \Else

            // Link cluster to low-voltage line segment
                \State $E_{Last} \leftarrow E_{Last} \cup \text{Path}(B_l^*, P_l^*, \mathcal{R})$ 
            \EndIf
        \EndFor
    \EndFor
    
    \Return $E_{Last}$
\end{algorithmic}
\end{algorithm}

The cluster is then linked to the target that provides the shortest distance $D=min(D_p,\; D_l)$. We will generate the link between the building that achieves this minimum distance and the corresponding nearest utility pole or a point on a distribution line. Note that the determined link should be generated following the road network topology. We present the full procedure of last-mile grid topology reconstruction in Algorithm~\ref{alg:last_mile_connectivity}.

The full power grid topology identification can be generalized by integrating the high- and medium-voltage network topology, the low-voltage assets connectivity, the intra-cluster building connectivity, and the cluster-to-skeleton links.

\section{Case-study: Power grid identification for Alna (Oslo, Norway)}\label{sec:case_study}

\subsection{Data collection and an overview of the studied area}

\begin{figure}[tbhp]
    \centering
    \includegraphics[width=0.9\linewidth]{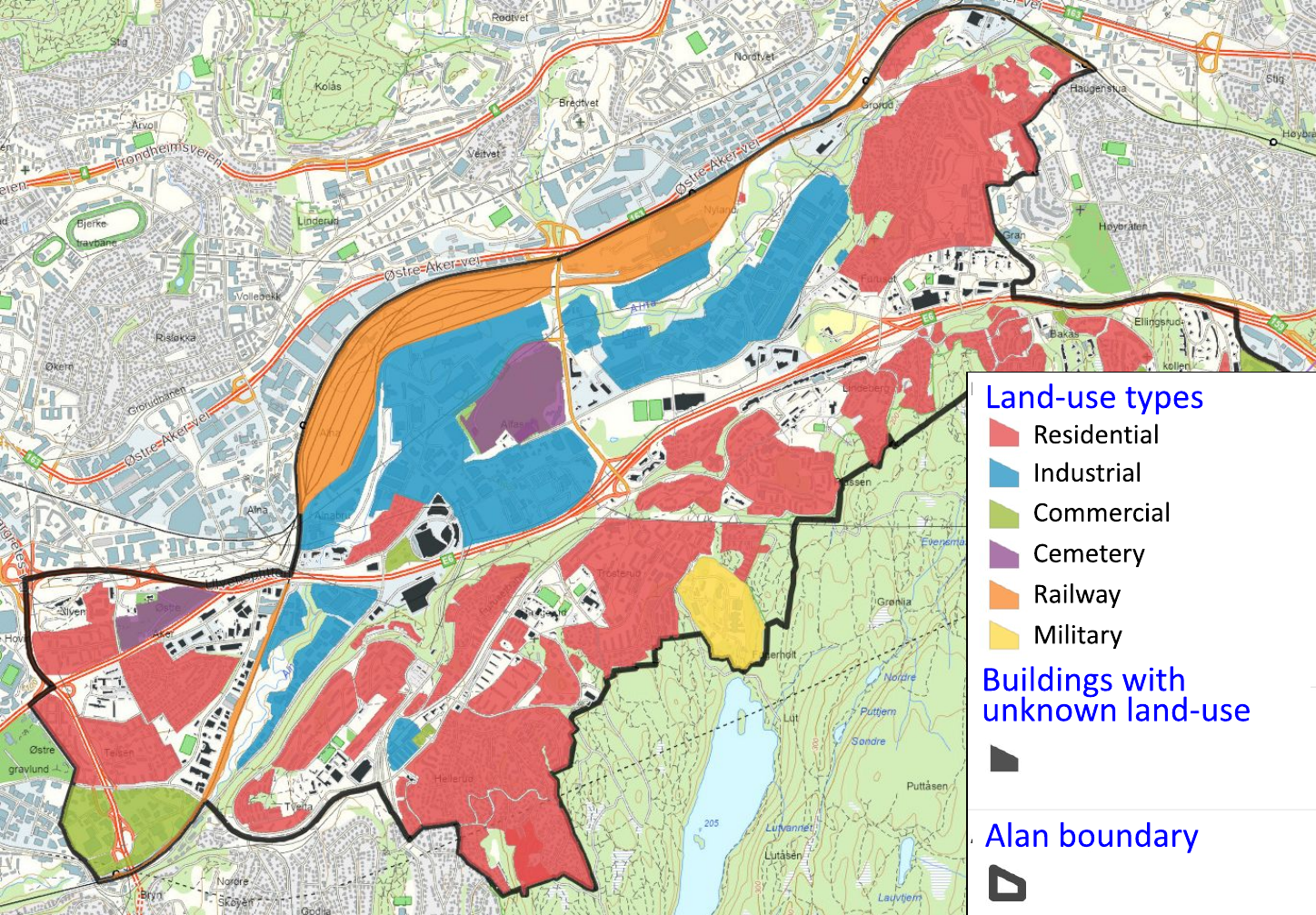}
    \caption{Land-use of different types in the area of Alna, Oslo.}
    \label{fig:alna_area}
\end{figure}

\begin{figure}[tbhp]
    \centering
    \includegraphics[width=0.9\linewidth]{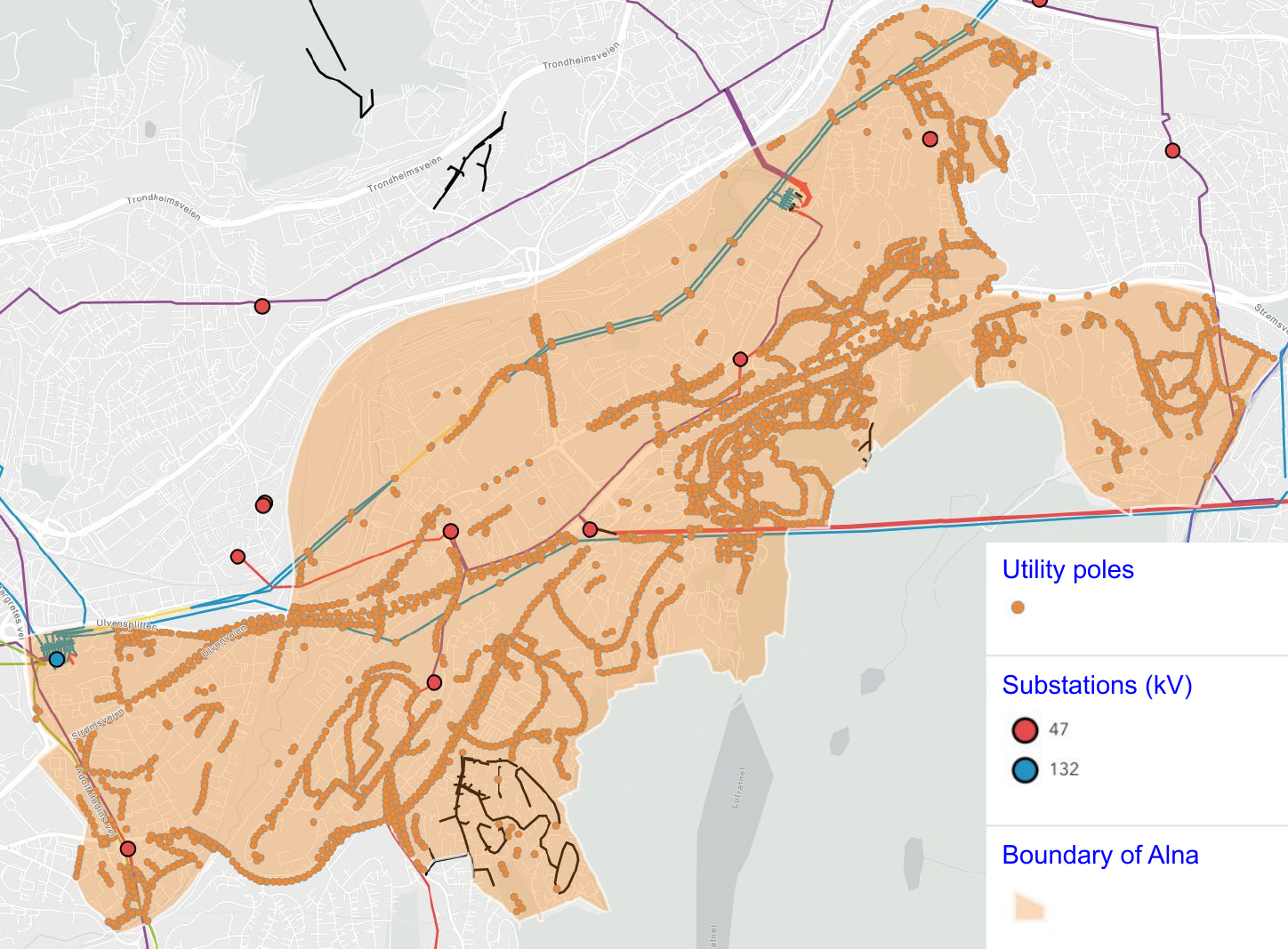}
    \caption{Utility assets in the area of Alna from the utility data released by NVE.no. The lines in different colors indicate power grids of different voltage levels.}
    \label{fig:alna_assets_osm_nve}
\end{figure}

\begin{figure}[tbhp]
    \centering
    \includegraphics[width=0.9\linewidth]{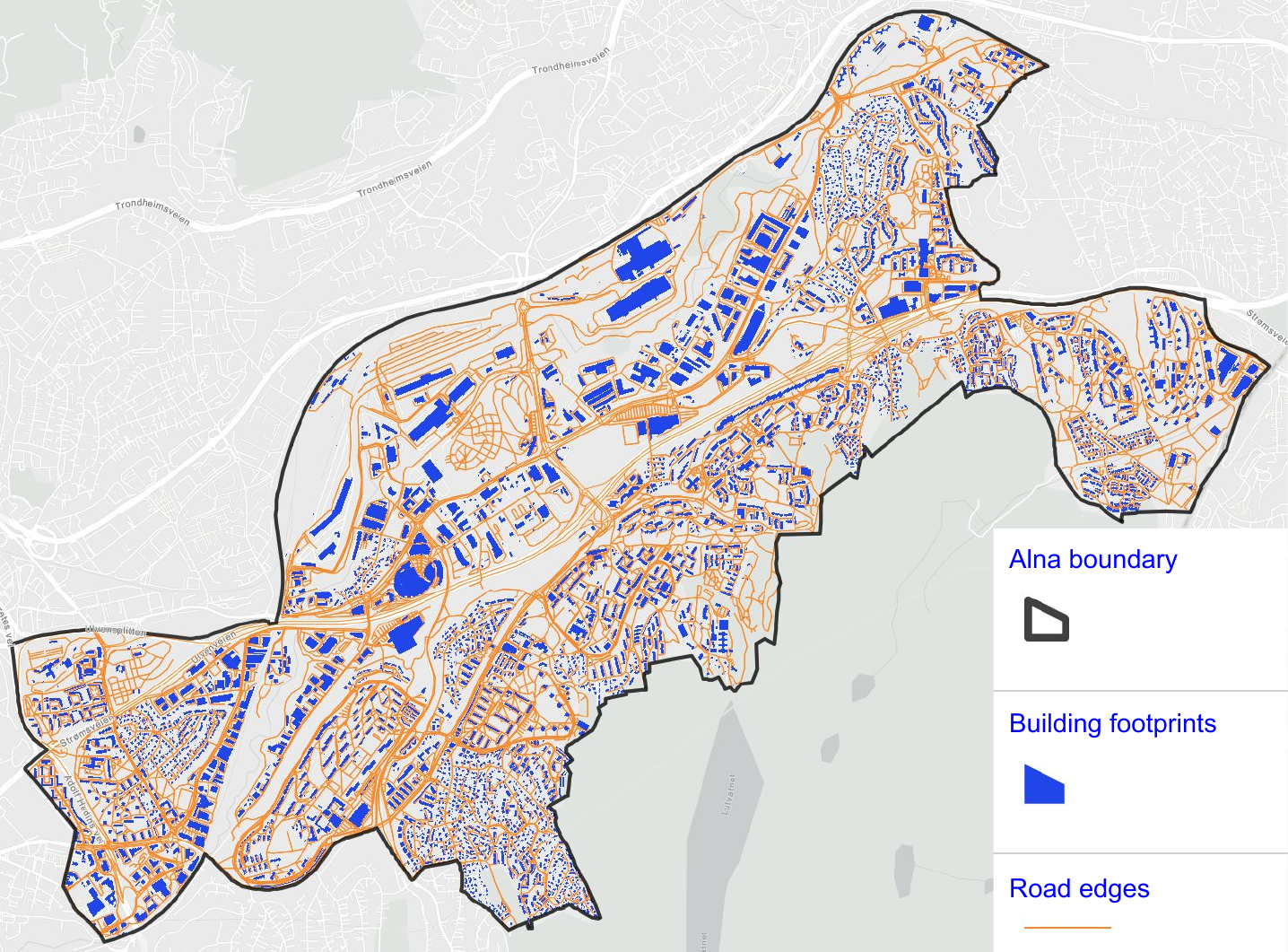}
    \caption{Building footprints and road networks in Alna.}
    \label{fig:alna_building_road_network}
\end{figure}

This section implements the reconstruction of the power grid network topology and hierarchy for the district of Alna, Oslo, Norway. Alna is selected as the case study area as it exhibits substantial diversity in land-use type (see Fig.~\ref{fig:alna_area}), power grid infrastructure (see Fig.~\ref{fig:alna_assets_osm_nve}), and a sufficient large number of buildings (see Fig.~\ref{fig:alna_building_road_network}). Table~\ref{tab:assets_overview} summarizes all existing power infrastructure assets extracted from OpenStreetMap and NVE, as well as the number of buildings and land-use types within Alna.

\begin{table}[tbhp]
    \centering
    \caption{An overview of power grid and infrastructure in Alna.}\label{tab:assets_overview}
    \resizebox{0.48\textwidth}{!}{
    \begin{tabular}{lc}
         \toprule
         Asset name & Total number of the asset in Alna\\
         \toprule
         Area in m$^2$ & 13,680,805.23\\
         \hline
         Buildings & 7,330\\
         \hline
         Types of land-use & \tabincell{l}{51 residential areas\\9 industrial areas\\7 commercial areas\\5 cemetery areas\\4 railway areas\\1 military area}\\
         \hline
         \tabincell{l}{Buildings without\\ land-use type} & 683 \\
         \hline
         Power lines & 127\\
         \hline
         Transformers & 8 substations\\
         \hline
         Utility poles & 2,787\\
         \hline
         Roads & 21,406 road edges\\
         \bottomrule
    \end{tabular}}
\end{table}

\subsection{Grid skeleton Identification}

\begin{figure}[tbhp]
    \centering
    \includegraphics[width=0.9\linewidth]{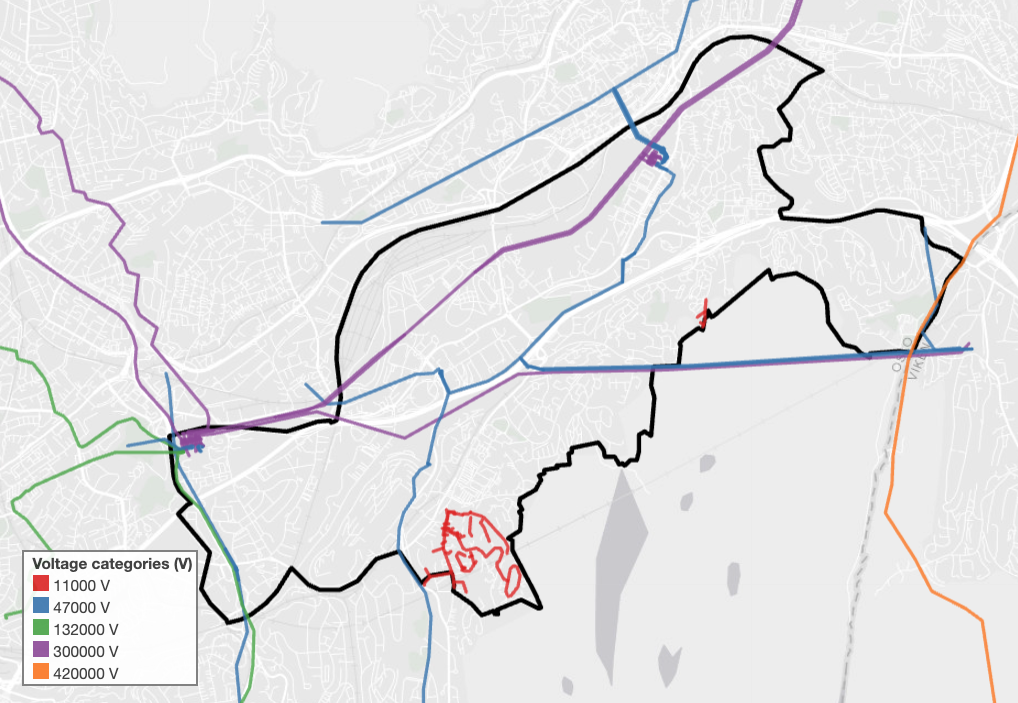}
    \caption{All the power grids GIS information integrated from OpenStreetMap and utility data from NVE.}
    \label{fig:alna_assets_osm_nve_grid_only}
\end{figure}

This section integrates the existing power grid data from OpenStreetMap and NVE, and connects power line segments that share common nodes. This procedure serves as a pre-processing step on the raw data. The outcome is a connected transmission and main distribution power grid network with five voltage levels, as shown in Fig.~\ref{fig:alna_assets_osm_nve_grid_only}. In total, we obtain 34 power lines at 11 kV, 59 lines at 47 kV, 3 lines at 132 kV, 30 lines at 300 kV, and 1 line at 420 kV. Although this subset of lines is not very dense, it constitutes the grid that links transmission system operator (TSO) to the local distribution system operator (DSO).

\subsection{Last-mile power grid network identification}

This section applies Algorithm~\ref{alg:1}, \ref{alg:lv_connectivity}, and \ref{alg:last_mile_connectivity} to identify the last-mile power grids that connect directly to individual buildings. We also determine how these last-mile connections link to utility poles and the low-voltage distribution network. 

\begin{figure}[tbhp]
    \centering
    \includegraphics[width=0.9\linewidth]{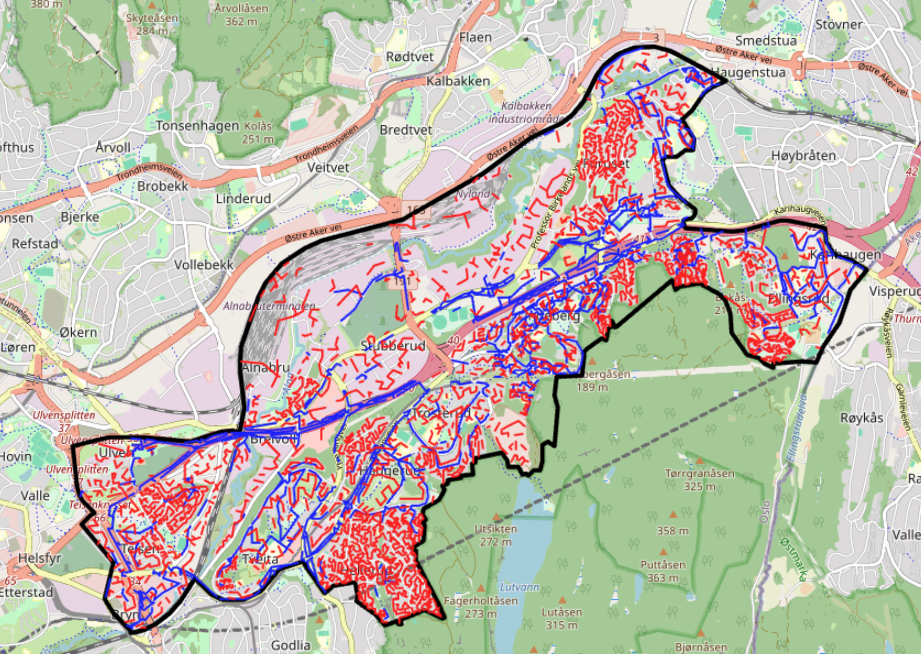}
    \caption{Identified last-mile power line connections between buildings (in red lines) and utility power poles (in blue lines).}
    \label{fig:pole_and_building_connections}
\end{figure}

\begin{figure*}[tbhp]
    \centering
    \includegraphics[width=0.99\linewidth]{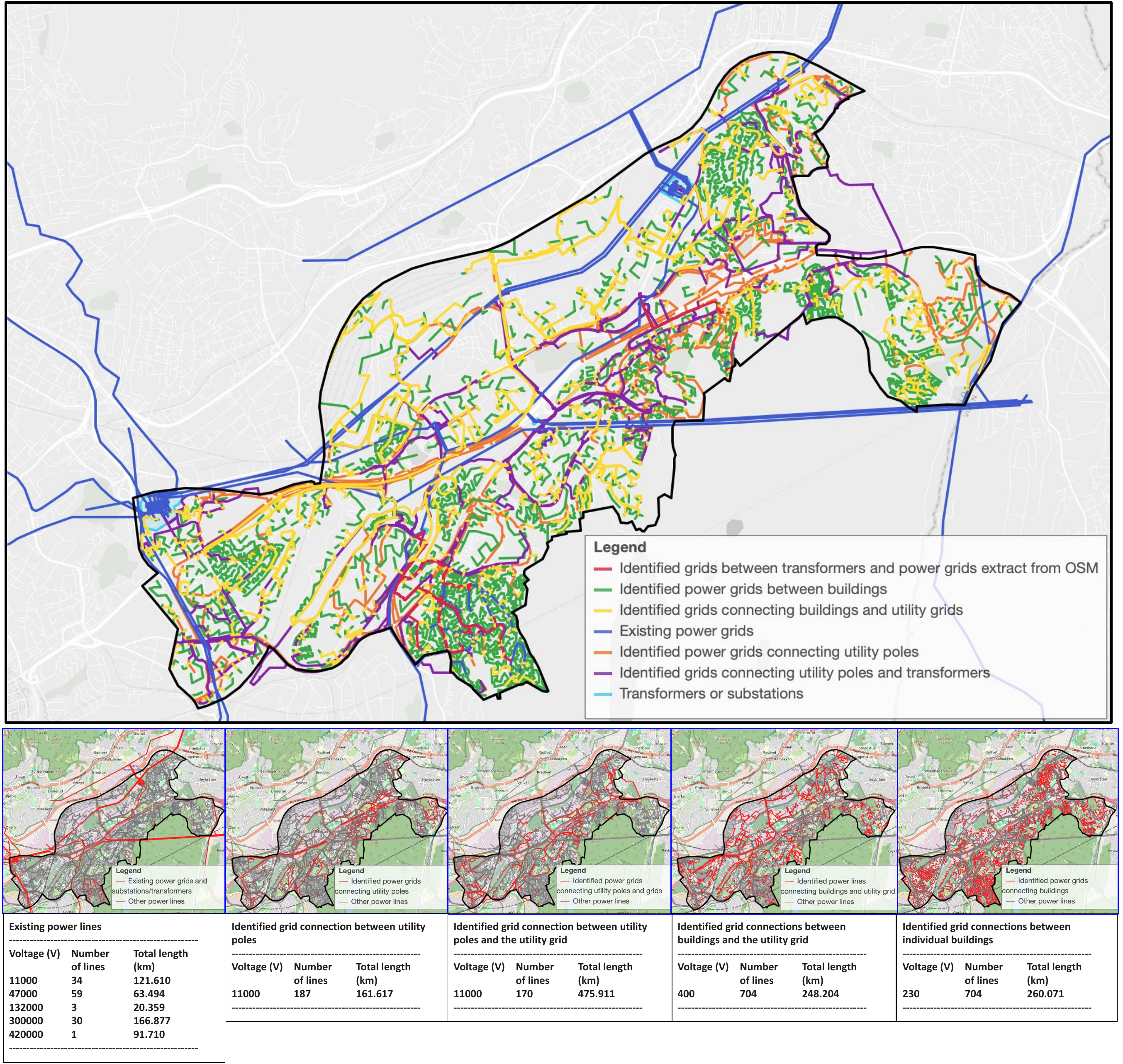}
    \caption{Overview of the identified power grids from the skeleton grid down to the last miles for Alna, Oslo.}
    \label{fig:final_identified_grid}
\end{figure*}

The identified power grids connecting utility poles and individual buildings are illustrated in Fig.~\ref{fig:pole_and_building_connections}. As expected from the design of Algorithm~\ref{alg:lv_connectivity}, the inferred utility pole connections generally follow the local road network. We summarize the numerical grid identification results in Fig.~\ref{fig:final_identified_grid}, where we visualize different identified grid components according to their position in the hierarchy. The figure shows (i) the existing grid topology and transformers (in the left) from OSM and NVE (ii) the identified power connections between utility poles (iii) the identified power connections between the main distribution grid, transformers, and utility poles (iv) the identified grid connecting individual buildings, and (v) the identified grid that connects buildings with utility poles and low-voltage distribution grids.

\begin{figure}[tbhp]
    \centering
    \includegraphics[width=0.74\linewidth]{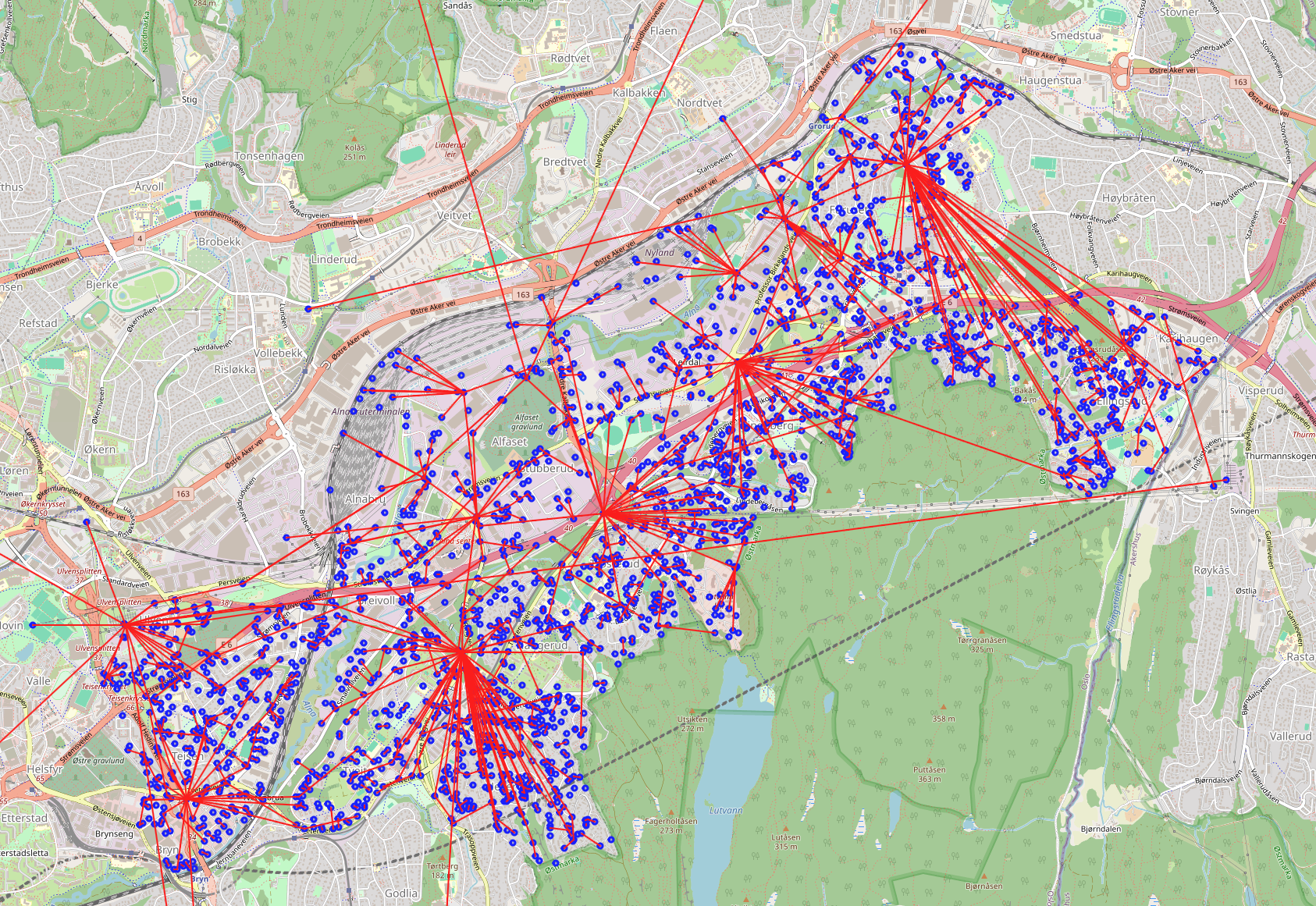}
    \caption{Simplified view of identified grids for Alna, with power lines straightened to visualize grid hierarchy. Blue dots denote buildings, and the red lines represent power lines between substation and the load.}
    \label{fig:bus_version}
\end{figure}

Fig.~\ref{fig:final_identified_grid} indicates that the power grids connecting utility poles and the distribution grids (with a length of 475.911 km in total) and those linking individual buildings and distribution grids (length of 248.204 + 260.071 km in total) account for the majority of the total network length in our identification. These components are typically absent from publicly available datasets. For clarity, we also provide a simplified view of the identified power grid for Alna in Fig.~\ref{fig:bus_version}, which offers an intuitive visualization of the grid hierarchy. This representation shows how individual building loads are connected and aggregated to their nearest substations or transformers, which are further linked to high-voltage distribution and transmission networks.

\section{A further case study on Waiheke Island, New Zealand}\label{sec:further_case_study}

We apply our approach in a new area, the island of Waiheke in New Zealand. It has a nearly close grid network with uneven distributions of cables and buildings. This case study aims to examine our approach in significantly different network scenario. We focus on the numerical results in this case.

\begin{figure*}[tbhp]
    \centering
    \includegraphics[width=0.99\linewidth]{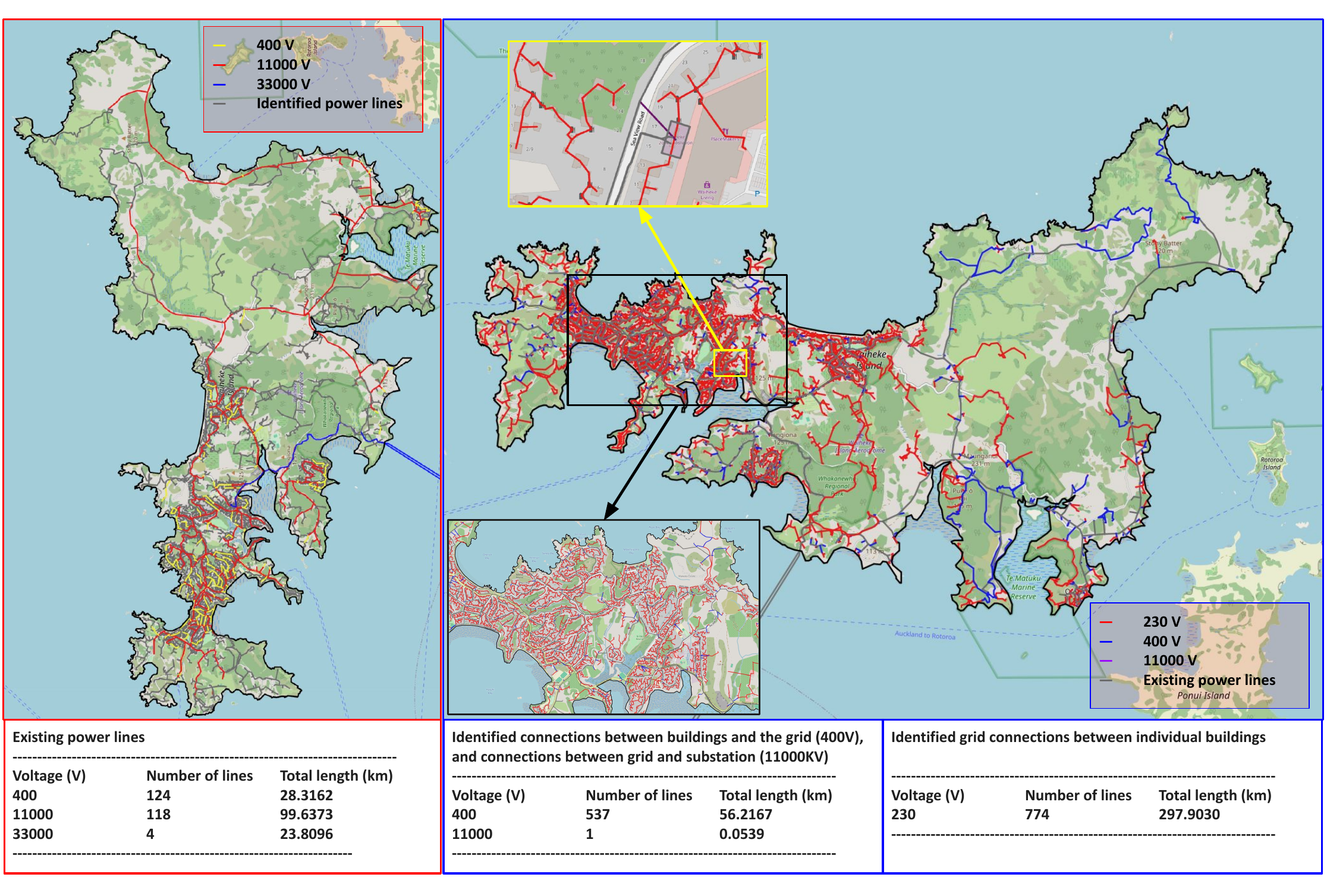}
    \caption{Overview of the existing and identified power grids for Waiheke.}
    \label{fig:final_identified_grid_Waiheke}
\end{figure*}

\begin{table}[tbhp]
    \centering
    \caption{An overview of infrastructure and landscape in Waiheke.}\label{tab:assets_overview_Waiheke}
    \resizebox{0.48\textwidth}{!}{
    \begin{tabular}{lc}
         \toprule
         Asset name & Total number of the asset in Waiheke\\
         \toprule
         Area in m$^2$ & 92,964,518.37\\
         \hline
         Number of buildings & 11,795\\
         \hline
         Types of land-use & \tabincell{l}{243 residential areas\\171 wood areas\\122 vineyard\\90 orchard\\88 scrub areas\\47 meadow areas\\11 industrial areas\\10 commercial areas\\7 heath areas\\4 cemetery areas}\\
         \hline
         \tabincell{l}{Buildings without\\ land-use type} & 2,001 \\
         \hline
         Power lines & 249\\
         \hline
         Transformers & 1 substation\\
         \hline
         Utility poles & 2,512\\
         \hline
         Roads & 9,609 road edges\\
         \bottomrule
    \end{tabular}}
\end{table}

We extract from OSM the existing power infrastructure and landscape in Waiheke, and list them in Table~\ref{tab:assets_overview_Waiheke}. Waiheke is a much larger area compared with Alna in Oslo, with more buildings and labeled power grids from OpenStreetMap, and similar number of utility poles. But it has only one  transformer and much less number of roads, and much more land-use types reflected by its island landscape nature. This leads to a very distinct scenario compared with the case of Alna. Since we only have one substation in Waiheke (33kV: 11kV) that is already connected with the existing 33KV power grid, we will connect all the 11kV power lines to that substation (see Algorithm~\ref{alg:1}). We assume that the voltage conversion between lower voltage levels (\textit{e.g.}, from 11kV to 400V) happens at pad-mounted and/or pole-mounted transformers, which are located at the intersections of power lines of different voltage levels. We also note that the utility poles are all connected with power lines from OpenStreetMap. Therefore, we did not execute the connecting of utility poles demonstrated in Algorithm~\ref{alg:lv_connectivity}.

We demonstrate the identified Waiheke power grids along with the existing grids from OSM in Fig.~\ref{fig:final_identified_grid_Waiheke}. The result shows that we have less types of identified grids compared with Alna. This is because the utility poles in Waiheke are all connected with power grids in OSM already, thus sparing the efforts in linking them with grids and buildings. We also note that Waiheke's power grid hierarchy is less complex compared with Alan. This is because Waiheke's building types are dominated by residential, while Alna has a higher portion of commercial and industry areas and it also has more substations to supply different types of building loads. It also shows that, while the total area differs a lot between Waiheke and Alna, the number of identified power lines connecting buildings is similar since they have similar number of buildings, which is as expected.

\section{Implications of the distribution network with the last-mile}\label{sec:implication}

The reconstruction of the last-mile power grids adds a critical component to the existing representations of the grid topology. It enables physics-based modeling of the network that links the granular, building-level demand to the utility grid. In our approach, each edge in the identified grid corresponds to an individual building indexed at OpenStreetMap. These buildings are, by their nature, associated with rich attributes such as building type and number of floors, and can be further linked to local power production (\textit{e.g.}, rooftop PV), population density, and local temperature profiles. Such information can support the construction of high-resolution energy demand profiles for individual buildings, leading to realistic load patterns on the identified grid network. This makes it possible to build an energy digital twin for urban areas that simultaneously respects the underlying power grid topology, geographic constraints, and fine-grained demand characteristics. 

At the same time, this level of detail substantially increases the computational burden of power flow analysis, as it introduces a large number of demand nodes. In our case-study for the Alna district in Oslo, we have more than 7,000 individual buildings as load points. Extending this approach to the entire Oslo city increases this number to over 80,000. This scale requires not only additional computational overhead for generating and maintaining building-level demand profiles, but also increased complexity for real-time power flow simulations. Aggregating the loads at selected points in the network can reduce the problem to a tractable size. However, the trade-off between granularity of the power flow analysis and the associated computational complexity requires careful consideration.

We also note that the identification of power grid hierarchy and topology is extensible, and once identified, the generated knowledge can be easily revised and calibrated further. This is because all the information and data used in the identification is georeferenced, so is the case with the identified grid topology and hierarchy. Thus, the generated topology can be inspected by GIS software platforms like QGIS or ArcGIS, and scrutinized via map services like Google Maps or Bing Maps. In this way, our method provides the grid identification that can be a basis for future studies to build upon on, and to extend the network identification in scale (\textit{i.e.}, for other regions) or in depth (\textit{e.g.}, for other power infrastructure and energy attributes).

\section{Conclusion and future research}\label{sec:conclusion}

This work presents an approach to identify urban power grid topology and the hierarchy while respecting geographical and physical constrains. The identified power grid connects the transmission grid down through the main distribution grid, low-voltage distribution grids, and ultimately to the last-mile connections to individual buildings. We develop the grid topology leveraging geographic machine learning on open data, ensuring that our method generalizes across regions. Our work opens opportunities to construct physics-based energy digital twins, to simulate the power flow with fine-grained demand profiles, and to assess grid resilience under increasing penetration of distributed energy production. 

It is important to note that 100\% accurate reconstruction of grid topology is impossible, particularly because the ground-truth data are missing for the last-mile connections. Nevertheless, a realistic topology and hierarchy, derived from existing power infrastructure and geographical constrains, can serve as a critical tool for analyzing power flow dynamics and gaining insights into grid behavior and the demand-supply balance.

An immediate future work is to synthesize or estimate large-scale building energy demand. This will enable approximation of realistic power load distributions, which are essential for the power flow analysis and simulation. We also suggest investigating the trade-off between fine-grained load representation and the computational complexity, in order to better understand the feasibility and scalability for creating energy digital twins for urban power systems.

\bibliographystyle{unsrt}
\bibliography{Reference}

\end{document}